\pdfoutput=1

\documentclass[11pt]{article}

\usepackage[final]{acl}

\usepackage{times}
\usepackage{latexsym}
\usepackage[T1]{fontenc}
\usepackage[utf8]{inputenc}
\usepackage{times}
\usepackage{latexsym}
\usepackage{hyperref}
\usepackage{url}
\usepackage{booktabs}
\usepackage{wrapfig}
\usepackage{graphicx}
\usepackage{enumitem}
\usepackage{color, colortbl}
\usepackage{caption}
\usepackage{subcaption}
\usepackage{multirow}
\usepackage{amsthm}	
\usepackage{amsmath}
\usepackage{algorithm}
\usepackage{algpseudocode}
\usepackage{titlesec}
\usepackage{float}
\usepackage{adjustbox}
\usepackage{wrapfig}
\usepackage{pifont} 
\usepackage{bm}     

\usepackage{microtype}

\usepackage{inconsolata}

\usepackage{xspace}
\newcommand{\ourapproach}{\texttt{Evolver}\xspace}
\title{Knowledge Fusion By Evolving Weights of Language Models}

\newcommand{\samethanks}[1][\value{footnote}]{\footnotemark[#1]}
\author{
Guodong Du$^{1}$ \quad
	Jing Li$^{1}$\thanks{\,\, Corresponding authors.} \quad
	Hanting Liu$^{2}$ \quad
	Runhua Jiang$^{2}$ \quad \\
	\textbf{Shuyang Yu}$^{2}$ \quad
	\textbf{Yifei Guo}$^{2}$ \quad
        \textbf{Sim Kuan Goh}$^{2}$\samethanks \quad
        \textbf{Ho-Kin Tang}$^{1}$\samethanks   \\
   $^{1}$Harbin Institute of Technology, Shenzhen, China\\
 	$^{2}$Xiamen University Malaysia \\
    \texttt{duguodong7@gmail.com} \quad \texttt{jingli.phd@hotmail.com} \quad \\ \texttt{simkuan.goh@xmu.edu.my} \quad \texttt{denghaojian@hit.edu.cn}
}

\begin{document}
\maketitle
\begin{abstract}
	
	Fine-tuning pre-trained language models, particularly large language models, demands extensive computing resources and can result in varying performance outcomes across different domains and datasets.
	This paper examines the approach of integrating multiple models from diverse training scenarios into a unified model. 
	This unified model excels across various data domains and exhibits the ability to generalize well on out-of-domain data. 
	We propose a knowledge fusion method named \ourapproach, inspired by evolutionary algorithms, which does not need further training or additional training data.
	Specifically, our method involves aggregating the weights of different language models into a population and subsequently generating offspring models through mutation and crossover operations. 
	These offspring models are then evaluated against their parents, allowing for the preservation of those models that show enhanced performance on development datasets.
	Importantly, our model evolving strategy can be seamlessly integrated with existing model merging frameworks, offering a versatile tool for model enhancement. 
Experimental results on mainstream language models (i.e., encoder-only, decoder-only, encoder-decoder) reveal that \ourapproach outperforms previous state-of-the-art models by large margins. The code 
is publicly available at \url{https://github.com/duguodong7/model-evolution}.

\end{abstract}

\section{Introduction}

Due to the high training costs of large language models, it is common practice to fine-tune already pre-trained language models to adapt them for specific applications. 
This fine-tuning approach often allows us to achieve excellent performance in specific data domains or tasks at a relatively lower cost \citep{chen2021transformer}. 
However, the challenge lies in the fact that fine-tuning \citep{dodge2020fine} the same model in different task scenarios may result in performance variations, meaning that the results may not be satisfactory when testing the same model in different contexts. 
Therefore, our objective is to integrate knowledge from models trained in different scenarios to enhance the model's performance in cross-domain or cross-task scenarios \citep{wortsman2022robust}, without the need for further training or extra training data.

Mainstream knowledge fusion methods can be divided into two main categories. The first involves extensive training on large datasets across multiple tasks to learn new model parameters with shared representations, such as in multi-task learning. The second relies on fusing existing models from specific scenarios without requiring extensive data. While multi-task learning generally improves overall performance, it has significant drawbacks: the need for abundant annotated data for all tasks and the complexity and time consumption of the training phase, especially with dataset combinations \citep{ruder2017overview}.
In contrast, model merging methods do not require retraining models and do not raise concerns about data privacy. In this paper, we primarily delve into the second category of methods and introduce an innovative model evolution approach inspired by Darwinian evolution theory \citep{shafiee2018deep}. 
In short, we compare our model evolution approach with other prevalent knowledge fusion methods, detailing their distinct features in Table~\ref{tab: comparison}.


\begin{table*}[t]
	\vspace{-1cm}
	\centering
	\scalebox{0.7}{
		\begin{tabular}{ccccccc}
			\toprule
			& Ensemble  & Model Merging & Multitask Learning & Federated Learning & Model Soups  & \textbf{Model Evolution (ours)}  \\
			\midrule
			Retraining & \ding{55} & \ding{55} & \ding{51} &\ding{51} & \ding{55} & \ding{55} \\
			High Memory Cost & \ding{51} & \ding{55} & \ding{55} & \ding{55} & \ding{55} & \ding{55}  \\
			Round(s) & Single & Single & Single & Multiple & Greedy & Greedy \\
			Data & No & A Few Examples &Train Datasets & Private & Dev Datasets & Dev Datasets \\
			Key Technique &  Inference & Matrices Computing & Distribution & Back-Propagation & Search &  Evolution \\
			Peak GPU Memory &\ding{55} & \ding{51}  & \ding{51} & \ding{51}  &\ding{55} & \ding{55}  \\
			\bottomrule
	\end{tabular}}
	\caption{Comparison of different knowledge fusion methods. Round means the number of times the models are edited when implementing a certain knowledge fusion method. Peak GPU memory is used to compare the GPU memory requirements.}
	\label{tab: comparison}
\end{table*}

In fact, the problem of model merging can be reformulated as an optimization problem that aims to find the most valuable information for knowledge fusion in order to achieve better outcomes than any individual model alone. For instance, \cite{jin2022dataless} employed a simple linear regression approach for optimization, while model soups \citep{wortsman2022model} implemented a greedy search method. In this paper, we consider the adoption of a more robust evolutionary algorithm for optimization. Evolutionary algorithms offer several advantages, including their outstanding performance in handling complex, high-dimensional, and nonlinear problems, as well as their relative insensitivity to local optima. 
Traditionally, evolutionary algorithms are primarily used in neural architecture search (NAS)~\citep{awad2020differential}. However, in this paper, we pioneer their application to the selection of important weights in language models for knowledge fusion.

\begin{figure}
	\centering
	\includegraphics[width=0.49\textwidth]{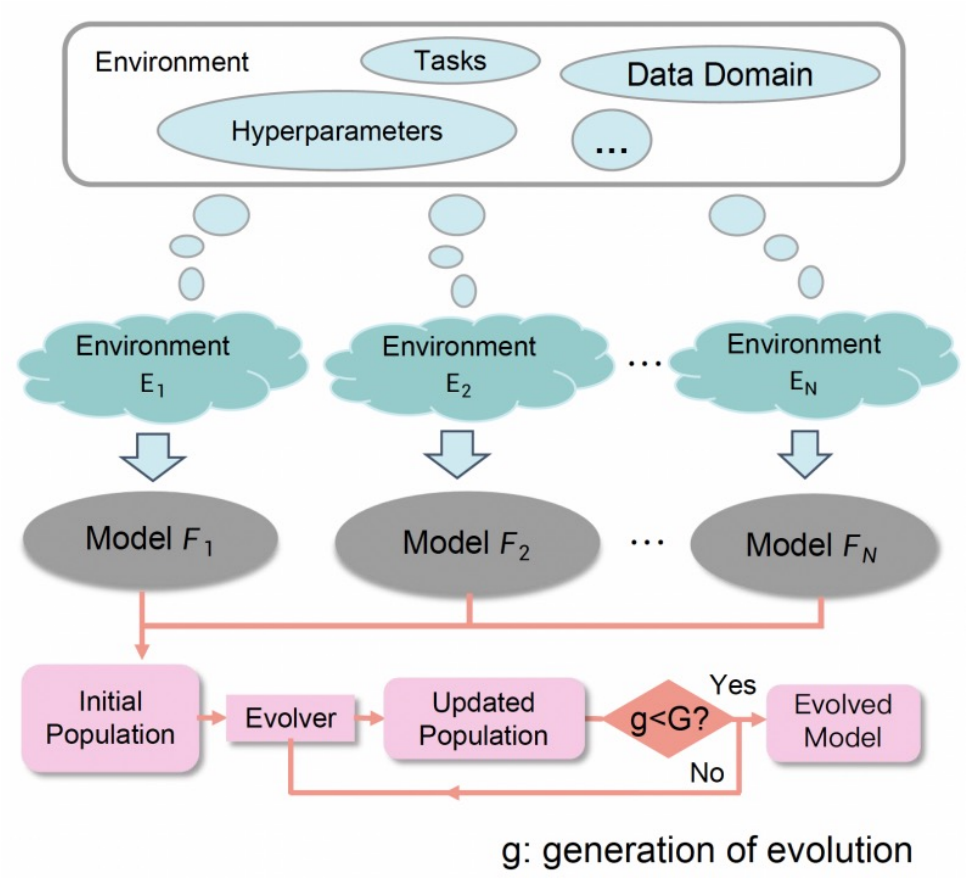}
	\caption{Key idea illustration. The key idea in our model evolution method is to aggregate multiple models $f_{1..N}$ from various environments into a population, which is then iteratively updated through greedy evolutionary rounds.}
	\label{fig:model-evolver-overview-intro}
	\vspace{-0.3cm}
\end{figure}

As shown in Figure~\ref{fig:model-evolver-overview-intro}, our approach first processes models fine-tuned in different environments as an initial population. We then generate a new population through mutations and recombination among different individuals within the population. 
Subsequently, we validate the performance of the new population on development environment datasets and preserve elite individuals for updating. After evolving with enough generations, we select individuals with the best performance as the evolved model. 
Our evolutionary algorithm is firmly rooted in the task vector \citep{ilharco2022editing} methodology, wherein we derive the differential vectors between two individuals during the mutation operation. We delve deeper into the task vector concept from two distinct perspectives. Firstly, we employ a scaling factor  ($F$), to search and regulate the weights of the difference vectors. This is crucial for aligning the expectations of model outputs, as illustrated in \citep{yu2023language}. Secondly, we utilize the crossover ratio ($Cr$), to maintain a certain level of sparsity, thereby preventing interference among different individuals, as demonstrated in \citep{ties}. Additionally, by conducting random searches for crossover parameters, we are able to obtain more optimal per-parameter coefficients for model merging.

We conduct knowledge fusion experiments across various difficulty levels, employing different types of models, such as RoBERTa, DeBERTa, T5 and GPT2. 
These experiments cover sentiment classification tasks in diverse data domains, as well as benchmark tasks from the GLUE dataset. 
The experimental results consistently demonstrate that our proposed method effectively enhances performance across all experimental settings. Furthermore, our approach can be synergistically combined with previous model merging methods (\textit{e.g.}, \textit{fisher} \citep{matena2022merging}, \textit{regmean} \citep{jin2022dataless}), resulting in further improvements in knowledge fusion performance. This combined approach significantly outperforms baseline methods and previous techniques. Notably, our method also exhibits superior generalization performance when applied to data domains beyond the scope of multiple datasets.
To summarize, our key contributions include:
\begin{itemize}[noitemsep,nolistsep]
	\item \textbf{Innovative model evolution method:} We propose a novel knowledge fusion method from an evolutionary perspective, by evolving weights of language models. 
	\item \textbf{Improved knowledge fusion performance:} Our method consistently enhances knowledge fusion performance across a broad spectrum of data domains and tasks.
	\item \textbf{Effective integration with existing model merging methods:} Our approach can be effectively enhanced and augmented through the integration of existing model merging techniques.
\end{itemize}

\begin{figure*}[t]
	\includegraphics[width=\textwidth]{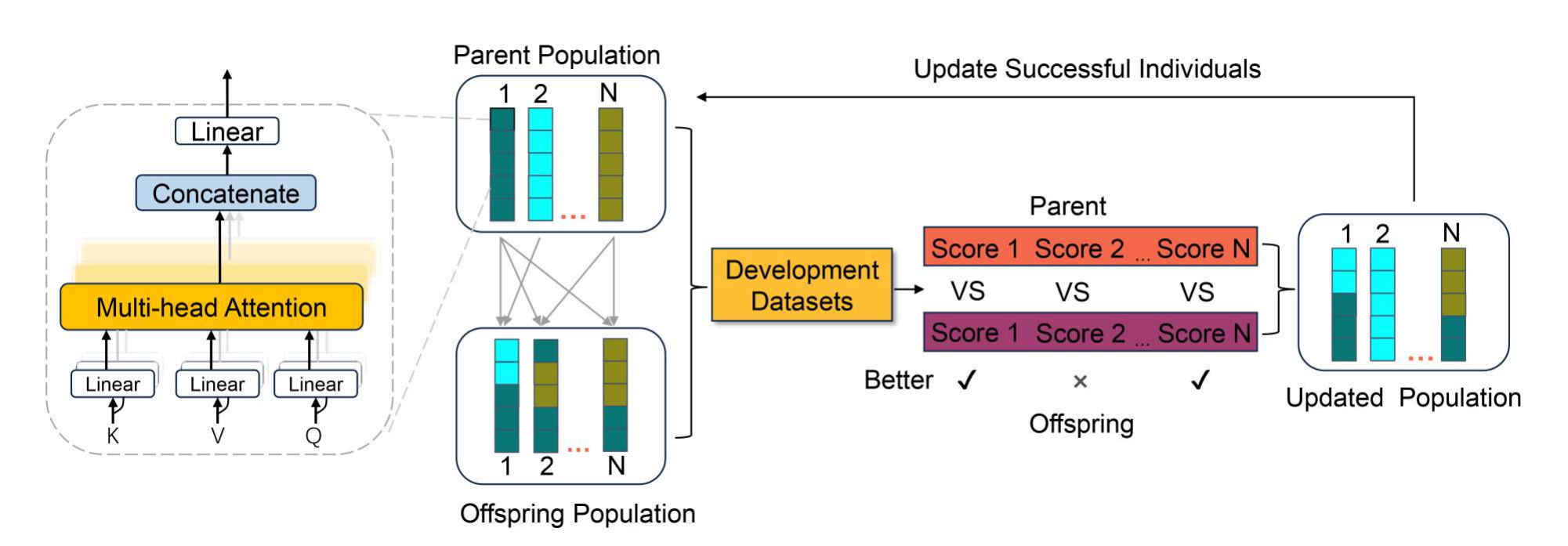}
	\vspace{-6mm}
	\caption{The process of evolving weights of language models. \ourapproach involves aggregating the weights of  language models into a population and  generating offspring models through mutation and crossover operations. }
	\label{fig:process of evolving}
	\vspace{-0.1cm}
\end{figure*}

\section{Related Work}
\subsection{Knowledge Fusion}
Numerous studies have shown that aggregating knowledge from multiple datasets can enhance the performance of a single model across various data domains and different tasks \citep{poth-etal-2021-pre}. This approach is also applicable to out-of-domain data \citep{wang2020multi}. \cite{frankle2020linear} demonstrated the effectiveness of simple weight averaging in model fusion, exhibiting better performance than pre-training methods. \cite{matena2022merging} proposed \textit{fisher-weighted averaging} to merge models with different architectures, taking into account the importance of each parameter. \cite{jin2022dataless} investigated model fusion using regression mean, re-weighting, and linearly combining rows within the weight matrix. \cite{wortsman2022model} introduced\textit{ greedy soup}, a technique to obtain robust results by searching for different average weights from multiple fine-tuned models. \cite{ilharco2022editing} proposed the concepts of task vectors to improve pre-trained models on multi-tasks and \cite{yadav2023ties} worked on the interference problems of task vectors to avoid the loss of valuable information. In addition to these knowledge fusion methods that do not require training, there are many knowledge fusion strategies that require complex training environments. Multi-task learning, as explored by \cite{ruder2017overview}, improves performance across various tasks by sharing knowledge. Federated learning \citep{wang2020federated} is a collaborative decentralized privacy-preserving technology designed to overcome the challenges of data silos and data sensitivity. 
Our evolutionary algorithm is firmly rooted in the task vector \citep{ilharco2022editing} methodology, wherein we derive the differential vectors between two individuals during the mutation operation. The methodology of task vectors has recently emerged as a promising approach for model merging and knowledge fusion, as evidenced by studies such as \citep{ties, ortiz2024task}. These investigations underscore the capacity of differential vectors, acquired through differentiation of pre-trained and fine-tuned models, to facilitate model editing. Overall, our approach harnesses the power of task vectors to enhance model merging and knowledge fusion, offering insights into both the theoretical and practical aspects of this methodology.

\subsection{Evolutionary algorithms}
Of particular relevance to our work is evolving algorithms (EAs), which provide an alternative path for addressing optimization problems in deep neural networks (DNNs) without relying on gradient information. The fundamental idea behind EAs is to combine the structures and weights of a group of neural networks and continually evolve them in the direction of improved global fitness to enhance model performance. These methods encompass genetic algorithms~\citep{montana1989training}, genetic programming~\citep{suganuma2017genetic}, differential evolution (DE)~\citep{pant2020differential, jiang2024cade}, and evolutionary strategies~\citep{salimans2017evolution}, among others. Neuro-evolution techniques, such as NEAT (Neuro Evolution of Augmenting Topologies)~\citep{stanley2002evolving}, have demonstrated the ability to design simpler neural network architectures for improved performance, particularly in reinforcement learning tasks. However, it's important to note that EA methods typically perform well on small datasets and small-scale DNNs~\citep{piotrowski2014differential}. When applied to large-scale datasets,
these methods tend to converge slowly and may even fail to converge~\citep{piotrowski2014differential}. 
In this paper, we propose model evolution which is motivated by the fact that model merging is neither amenable to traditional gradient-based optimization methods, nor are simple techniques like grid search sufficient. Therefore, we turn to evolutionary algorithms, which show promise for effectively addressing the model fusion problem.

\section{Evolving Weights of  Language Models}
The goal of model evolution is to combine multiple fine-tuned language models into a more powerful single model. We achieve this by simulating the evolution process of a neural network population, as shown in Figure~\ref{fig:process of evolving}. We use the same pre-trained checkpoint and fine-tune it in different environments to create the initial population. As all individuals share the same model architecture, this enables our evolution algorithm to perform mutations and recombinations among individuals within the parameter space. 

\subsection{Evolutionary Strategy: \ourapproach}

		\paragraph{Population Initialization.} 
		For the optimization problem of model merging, an original set of individuals (population) is initialized. The parameters of each of N models are flattened into a one-dimensional vector, forming a set of candidate solutions. In this way, we obtain a set of candidate individuals represented by $\theta={\theta_i}, i=1, ..., N$. Here, $N$ denotes the size of the population, and $\theta_i = (\theta_{i,1}, \theta_{i,2}, ..., \theta_{i,d})$ represents each candidate individual, where $d$ is the dimension. 
		
		\paragraph{Evolution Process.} 
		We simulate the evolution process of a population of neural networks using the differential evolution algorithm~\citep{pant2020differential}. Each generation consists of three main steps: mutation, crossover, and updating.
		
		\paragraph{Mutation:~}
		For each candidate individual $\theta_i$, we randomly select two other candidate individual $\theta_{r_{1}}$ and $\theta_{r_{2}}$, where $r_{1}$ and $r_{2}$ are two distinct random integers less than or equal to $m$. We use a scaling factor $F$ to adjust the differences between $\theta_{r_{1}}$ and $\theta_{r_{2}}$, and then add them to $\theta_{i}$ to obtain the mutated solution $\theta^{\star}_i = \theta_i + F\times(\theta_{r_{1}}-\theta_{r_{2}})$, where $F$ is used to control the weights of the difference vector in the new parameter set. 
		
		\paragraph{Crossover:~} The computation for crossover is as follows:
		\begin{eqnarray}
			\theta^{\star}_{i,j}  = \begin{cases}
				\theta^{\star}_{i,j} & \mbox{if} \quad \mbox{rand}(0,1)\leq Cr,\\
				\theta_{i,j} & \mbox{otherwise}.
			\end{cases}
		\end{eqnarray}
		where $Cr$ is the pre-set crossover degree threshold between the new individual and the parent solution, and the setting of the threshold $Cr$ can impact the ratio of elements selected in a mutated solution. 
		
		\paragraph{Updating:~}
		Throughout the process, we convert the offspring population vectors into models and conduct inference to get performance scores for these models on the development dataset. As demonstrated in the equation below, We sequentially evaluate the performance scores of offspring individuals in comparison to their parent one by one. If an offspring performs better, we then replace the corresponding parent individual with it, thereby updating the parent population.
		\begin{eqnarray}
			\theta_i = \begin{cases}
				\theta^{\star}_{i} & \textit{score}(\bm{\theta}^{\star g}_{i}) > \textit{score}(\bm{\theta}^{g}_i) \\
				\theta_{i} & \mbox{otherwise}.
			\end{cases}
		\end{eqnarray}

            \subsection{Integration with Other Merging Methods.}
            Our proposed model evolution method can be integrated with any other model merging technique. Specifically, we can select the optimal individual from the updated population as the outcome of the evolution process, which we refer to as a simple \ourapproach. Furthermore, we can also apply other model merging techniques to the updated population as a further improvement measure. Such integration can be implemented both at the stage of obtaining the final evolved model and during the calculation of the updated population's scores for iterative updates. This process has been summarized as Algorithm~\ref{algo:intergration} and is detailed in the flowchart provided in Appendix~\ref{appendix A}.

            \begin{algorithm}[t]
            \begin{minipage}{0.48\textwidth}
            \caption{Integrating Model Evolver with Other Model Merging Methods}
            \begin{algorithmic}[1]
            \State \hspace{0.0cm} \textbf{When} inference with the updadted population
            \State \hspace{0.2cm} \textbf{if} this is a simple evolver  
            \State \hspace{0.2cm} \textbf{then}
            \State \hspace{0.4cm} evaluate the performance of individual $\theta^\star_i$,
            \State \hspace{0.2cm} \textbf{else} 
            \State \hspace{0.4cm} merging $\theta^\star_i$ with other individuals with a specific merging method,
            \State \hspace{0.4cm} evaluate the performance of merged model.
            \end{algorithmic}
            \label{algo:intergration}
            \end{minipage}
            \end{algorithm}
 
		\subsection{Computation Efficiency}
		
		\paragraph{Memory Analysis:} The memory expanse during our model evolution is mainly related to the size of the population: $\sum_{i=1}^{N} d$, where $N$ represents the number of populations, $d$ is the dimension of the model parameter space. Since we avoid inner product matrices computing as in previous model merging methods such as \textit{fisher} and\textit{ regmean}, and the parameters is updated mainly through forward propagation of greedy models, the peak GPU memory consumption is consequently lower.
		
		\paragraph{Time Consumption:} We hereby provide the formula and key definitions required to calculate the runtime. The total evolving time can be calculated as $T = G\times N\times (t_1 + L\times t_2) \approx G \times N \times L \times t_2$, where $t_1 \ll t_2$ in practice. Here, $G$ is the total generations for updating, $N$ is the population size, $t_1$ is the time for mutation and crossover for each individual, $L$ is the number of samples of development datasets, $t_2$ is the time for inference of a sample on one model.
		
	
\section{Experimental Setup}

\subsection{Evaluation Settings}

\paragraph{Problems.} We primarily consider the following three main advantages when testing our proposed model evolution method: Firstly, we anticipate that our evolved model $f_M$, created by integrating knowledge from individual models $f_{1..N}$ finetuned on diverse datasets $D_{1..N}$, will have competitive performance across various data sources without necessitating separate models for each domain or task. Then, by evolving different models excelling in various tasks $D^{t}_{1..N_t}$, we aim to enhance multi-task handling capacity, avoiding the complexity of retraining as in MTL, while enabling cross-task inferencing within a single model. Lastly, our goal is for the evolved model $f_M$ to excel in generalizing to out of distributio (OOD) test sets $D^{o}_{1..N_o}$, thereby enabling it to effectively handle new and unforeseen data from domains or tasks not encountered during training. $D_{1..N}$.

\paragraph{Datasets.} We use the GLUE datasets~\citep{wang2018glue} as the cornerstone of our investigation into the performance of evolved models. This inquiry encompasses two key dimensions: training for non-independent and identically distributed (non-i.i.d.) partitions and training for disparate tasks. 
Detailed dataset information and additional experimental results are available in Appendix~\ref{appendix C}.

\paragraph{Implementation.}
We use Hugging Face's transformer library~\citep{Wolf2019HuggingFacesTS} to access pre-trained language models and conduct fine-tuning. All our models, denoted as $f_i$, follow the same architecture and employ identical pre-trained model weights $\theta$ for initialization, as described in ~\cite{mcmahan2017communication}. Our experiments include various pre-trained models as starting points, such as RoBERTa-base~\citep{liu2019roberta}, the lightweight DistilBert~\citep{khanuja2021mergedistill} and well-established model DeBERTa-large-v3~\citep{he2021debertav3}. Besides the models with encoder-only architecture,  we also conduct experiments with encoder-decoder architecture, T5-base-v1.1 ~\citep{t5} and decoder-only architecture, GPT2 ~\citep{gpt2} and large language model MiniCPM ~\citep{minicpm2024}.

\paragraph{Population Initialization.}
The initial population for model evolution is created through fine-tuning a model with the same initialization but on different data domains or various tasks. While fine-tuning DistilBERT-base, RoBERTa-base, and DeBERTa-large, we maintained a constant initial learning rate of 1e-5. Throughout our experiments, we consistently utilized the AdamW optimizer with a warm up learning rate during the initial 6\% of training. Our model training utilized a batch size of 16 and encompassed 10 epochs for the GLUE task and 30 epochs for the emotion classification task.
\subsection{Compared Methods}

We mainly compare \ourapproach with existing merging methods, including \textit{simple}, \textit{fisher}~\cite{matena2022merging} , \textit{regmean}~\cite{jin2022dataless} ,  \textit{greedy soup} ~\cite{wortsman2022model} and \textit{TIES}~\cite{yadav2023ties}. To gain a better grasp of the advantages of model merging, we show the performance 
prior to model evolution, the average performance of the population (\textit{Avg.$f_{1..N}$}) and the best-performing individual (\textit{Best.$f_{1..N}$}), more details is shown in Appendix ~\ref{app_c.2}. Moreover, we provide the performance for the model trained on a specific task \textit{domain-specific}. We also compare with \textit{model ensembling}, where the logits from predictions are extracted, averaged, and then subjected to the argmax operation. 
Lastly, we use multi-task learning (MTL) as a benchmark. 



\begin{table*}[t]
	\centering
	\scalebox{0.73}{
		\begin{tabular}{@{}ccccccc@{}}
			\toprule
			& \multicolumn{1}{c}{Encoder-Decoder} & \multicolumn{3}{c}{Encoder-only} & \multicolumn{2}{c}{Decoder-only} \\
			\cmidrule(r){1-1} \cmidrule(r){2-2} \cmidrule(lr){3-5} \cmidrule(lr){6-7}
			Method & T5-base &
			\begin{tabular}[c]{@{}c@{}}RoBERTa-base \\ Same / Diff Head Init.\end{tabular} &
			\begin{tabular}[c]{@{}c@{}}DistilBERT-base \\ Same / Diff Head Init.\end{tabular} &
			\begin{tabular}[c]{@{}c@{}}DeBERTa-large \\ Same / Diff Head Init.\end{tabular} & 
			GPT2 & MiniCPM \\
			\cmidrule(r){1-1} \cmidrule(lr){2-2} \cmidrule(lr){3-5} \cmidrule(lr){6-7}
			Avg. $f_{1..\textsc{N}}$  & 32.07 & 26.08  & 24.55 & 27.68 & 23.35 & 35.12 \\
			Best. $f_{1..\textsc{N}}$ & 34.08 & 29.27  & 29.91 & 31.93 & 26.76 & 37.23 \\
			Ensemble        & 33.95 & 38.77 / 27.73 & 26.51 / 25.43 & 29.88 / 29.27 & 26.82 & 37.05 \\
			Greedy Soup     & 34.10 & 30.34  & 30.11  & 31.93 & 26.76 & 37.31 \\
			\midrule
			Simple          & 39.47 & 23.18  & 23.70  &  3.75 & 21.54 & 42.44 \\
			\ourapproach (ours)         & 41.25 & 33.27 / 30.04 & 28.95 / 26.29 & 23.90 / 21.55 & 23.41 & 44.76 \\
			\midrule
			Fisher          & 39.12 & 26.09 / 22.43 & 26.39 / 22.61 & 12.83 / 20.42 & 24.93 & \textbackslash \\
			Fisher\_\ourapproach (ours) & 40.36 & 28.41 / 25.71 & 27.63 / 24.75 & 17.22 / 22.95 & 25.66 & \textbackslash \\
			\midrule
			RegMean         & 40.24 & 38.74 / 32.58 & 33.37 / 28.29 & 38.33 / 18.92 & 30.14 & \textbackslash \\
			RegMean\_\ourapproach (ours)  & 41.83 & 39.87 / 34.28  & 35.67 / 31.11 & 39.58 / 21.79 & 32.26 & \textbackslash \\
			\midrule
			TIES            & 41.24 & 39.66 / 35.32  & 35.55 / 30.14 & 39.22 / 21.67 & 32.11 & 45.81 \\
			\textbf{TIES\_\ourapproach} (ours) & \textbf{43.16} & \textbf{41.33} / \textbf{37.42}  & \textbf{37.12} / \textbf{31.48} & \textbf{40.61} / \textbf{23.24} & \textbf{33.56} & \textbf{46.24} \\
			\midrule
			Domain-Specific & 49.31 & 51.38         & 48.79         & 52.81   & 47.62 & 54.32 \\
			MTL             & 48.98 & 47.73         & 45.23         & 51.77   & 44.31 & 52.14 \\
			\bottomrule
		\end{tabular}
	}
	\caption{\small{In-domain performance when merging emotion classification models. The initial population are all 5 domain specific models or pairwise models. \textbackslash ~indicates the original merging methods can not been conducted due to high GPU cost. \textbf{Bold} numbers indicate the best performance across different model merging algorithms. All the results we reported are averages of trials conducted with 5 different random seeds.}}
	\label{tab:merge_all_in_domain}
 \vspace{-0.3cm}
\end{table*}

\section{Experimental Results}


We assess the performance dynamics of the model evolution method across a range of scenarios with varying levels of complexity. These scenarios include:
(1) performance across different data domains used for fine-tuning individual models.
(2) performance across different tasks, when individual models are specialized in only one task. (3) OOD generalization performance on datasets from previously unseen domains. 

\subsection{Model Evolving Across Data Domains}
\paragraph{Evolving All Domain-Specific Models.}\label{sec:5.1.1}

We conduct experiments of evolving five domain-specific models for emotion classification, and the results are recorded in Table \ref{tab:merge_all_in_domain}. 
Multi-task learning (MTL) achieves performance similar to that of domain-specific models, suggesting that a single model has the capability to acquire knowledge from multiple domains. Besides, \textit{model soup }approach, which greedily selects fusion objects, leads to some improvements over the best individual.
However, these improvements are relatively marginal compared to model merging methods.

We compare model evolution with three other knowledge merging methods. The basic version of model evolution outperformed \textit{fisher} method and achieved performance that is comparable to \textit{regmean} on some tasks. Furthermore, we explore the combined use of model evolution and model merging methods, demonstrating that our approach further enhances existing model merging methods and consistently yields improvements across different models.  Also, we demonstrate results with shared and different classification head initialization (Same Head Init/Diff Head Init). It can be observed that \textit{fisher} and \textit{regmean} produce unstable and highly variable results with different initialization, while the performance of the model evolution method is less affected by this factor. Therefore, our proposed model evolution method shows more stable performance when deploying and maintaining a single model across multiple domains.

\paragraph{Evolving Pairwise Domain-Specific Models}
In addition to the fusion experiment involving all the models, we also consider pairwise domain-specific models for knowledge fusion in the context of the emotion classification task. After pairing models from a set of 5, we conduct 10 ($\mathcal{C}^2_5$) runs with all the same methods described in section \ref{sec:5.1.1}. The result of these 10 runs are averaged and recorded in Table~\ref{tab:pairwise_emotion}. We observe clear differences between model evolution and model merging methods, with the \textit{TIES-evolver} achieving the best performance when combining pairwise models finetuned from different domains.
\begin{table}[htbp]
	\centering
	\scalebox{0.75}{
		\begin{tabular}{@{}c|cc|cc}
			\toprule
			Model & Simple & \ourapproach & TIES & \textbf{TIES\_\ourapproach} \\
			\cmidrule(r){1-1} \cmidrule(lr){2-3} \cmidrule(lr){4-5}  
			T5-base          & 38.82  & 40.21  & 46.35  & \textbf{47.92}  \\
			RoBERTa-base     & 37.78  & 39.13  & 45.56  & \textbf{46.89}  \\
			DistillBERT-base & 36.76  & 38.85  & 42.09  & \textbf{43.22}  \\
			DeBERT-large     & 38.11  & 39.46  & 45.87  & \textbf{46.78}  \\
			GPT2             & 36.33  & 38.25  & 41.85  & \textbf{42.62}  \\
			MiniCPM          & 40.22  & 42.71  & 47.02  & \textbf{48.83}  \\
			\bottomrule
		\end{tabular}
	}
	\caption{\small{In-domain performance when merging pairwise domain-specific emotion classification models. All the results we reported are averages of 10 ($\mathcal{C}^2_5$) runs after paring models from a set of 5.}}
	\label{tab:pairwise_emotion}
\end{table}

\begin{figure}[htbp]
	\begin{minipage}[b]{0.48\textwidth}
		\centering
		\includegraphics[width=\textwidth]{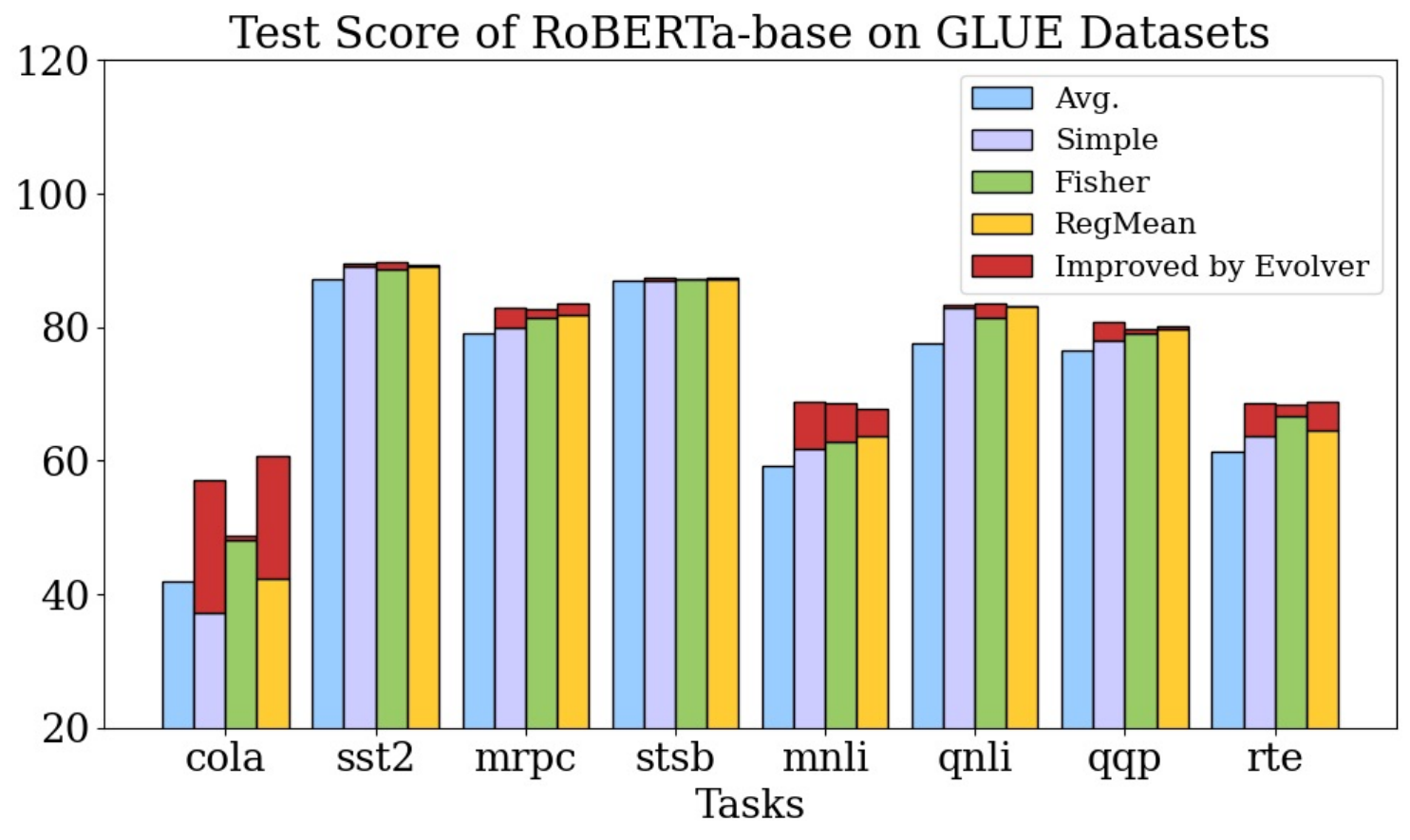}
		\caption{Result of model evolution on non-i.i.d. partitions of GLUE benchmark datasets.}
		\label{fig3:glue-niid}
	\end{minipage}
	\begin{minipage}[b]{0.48\textwidth}
		\centering
		\includegraphics[width=\textwidth]{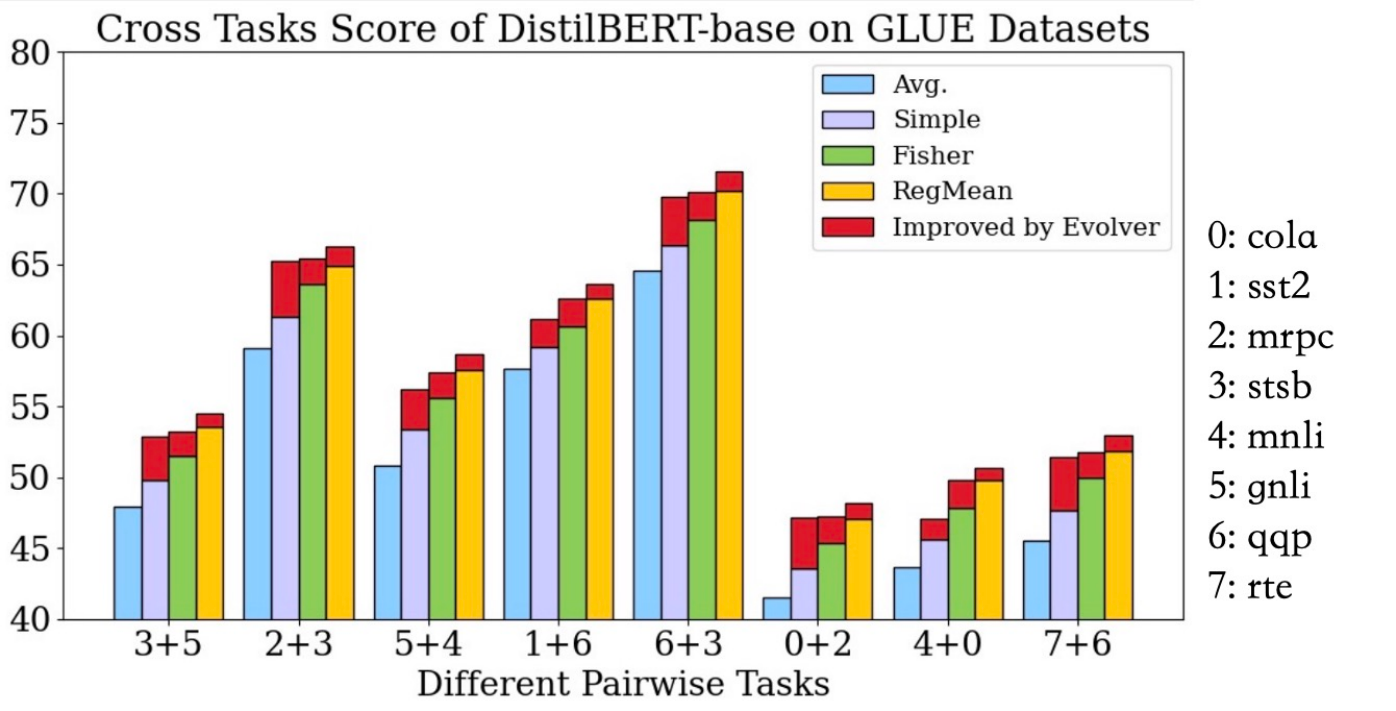}
		\caption{Result of model evolution across different pairwise tasks on GLUE benchmark.}
		\label{fig4:glue-diff-task} 
	\end{minipage}
\end{figure}

\paragraph{Evolving Models Trained on Non-i.i.d. Partitions.}
We adopt synthetic data divisions to simulate non-independent and identically distributed (non-i.i.d.) partitions of the same dataset, across the 8 tasks included in the GLUE benchmark. Given the inconsistency in the performance of \textit{regmean} and \textit{fisher} methods under different seeds, we choose to average the result of eight different random seeds. The results in Figure~\ref{fig3:glue-niid} indicate that the implementation of model evolution outperformed previous methods on all tasks, with clear improvements observed particularly on \textit{cola} and \textit{mnli} datasets. Details of this section are shown in Appendix ~\ref{app_c.1}.

\subsection{Evolving Models Across Different Tasks}
Here we examine the effectiveness of model evolution in merging models finetuned on different tasks. We use the RoBERTa-base model and train individual models with the complete training data for each task in the GLUE benchmark. Following this, we randomly select two task-biased individuals to conduct pairwise model evolution. Specifically, we exclude the parameters in task-specific headers due to their potential dimension variance depending on tasks. We summarize the results of eight different task pairs in Figure~\ref{fig4:glue-diff-task}, which show that our model evolution strategy performs effectively when fusing knowledge from diverse tasks.
\subsection{Model Evolving for Out-of-Domain Generalization}
For Out-of-Domain (OOD) generalization, we can obtain the same conclusion as our in-domain experiments, indicating that model evolving leads to improvements in OOD generalization performance, as summarized in Table~\ref{tab:merge_all_ood_emotion}. 
We notice that in the case of RoBERTa-base and DistilBERT models initialized with different heads, the basic \ourapproach has outperformed \textit{fisher-evolver} and \textit{regmean-evolver}. 
A plausible explanation for this is that previous methods may suffer from the negative impact of extremely poor-performing individual models. 
In contrast, model evolution possesses an elimination mechanism that effectively removes poorly-performed individual models during the competition process. 
This ensures that only models with superior performance are retained and merged. Consequently, model evolution can reduce the negative impact of under-performing models on the merged results.

\begin{table*}[htbp]
	\centering
	\scalebox{0.72}{
		\begin{tabular}{@{}ccccccc@{}}
			\toprule
			& \multicolumn{1}{c}{Encoder-Decoder} & \multicolumn{3}{c}{Encoder-only} & \multicolumn{2}{c}{Decoder-only} \\
			\cmidrule(r){1-1} \cmidrule(lr){2-2} \cmidrule(lr){3-5} \cmidrule(lr){6-7} 
			Method & T5-base &
			\begin{tabular}[c]{@{}c@{}}RoBERTa-base \\ Same / Diff Head Init.\end{tabular} &
			\begin{tabular}[c]{@{}c@{}}DistilBERT-base \\ Same / Diff Head Init.\end{tabular} &
			\begin{tabular}[c]{@{}c@{}}DeBERTa-large \\ Same / Diff Head Init.\end{tabular} & 
			GPT2 & MiniCPM \\
			\cmidrule(r){1-1} \cmidrule(lr){2-2} \cmidrule(lr){3-5} \cmidrule(lr){6-7} 
			Avg. $f_{1..\textsc{N}}$  & 30.12 & 20.92     & 19.69        & 21.17   & 18.63 & 35.24    \\
			Best. $f_{1..\textsc{N}}$ & 37.41 & 29.46     & 29.55        & 31.07   & 27.88 & 38.93   \\
			Ensemble         & 27.92 & 11.36 / 10.90 & 9.60 / 9.19   & 11.09 / 9.26   & 8.77 & 29.54 \\ 
			Greedy Soup      & 15.42   & 13.43     & 15.26        & 4.67    & 11.61 & 31.70    \\ \midrule
			Simple           & 38.61   & 11.56   & 13.21   &  0.24  & 10.25 & 40.21  \\
			\ourapproach (ours)         & 39.26   & 17.53 / 17.16 & 19.02 / 18.42 &  13.33 / 12.78  & 15.87 & 42.48 \\ \midrule
			Fisher           & 37.72   & 16.21 / 14.28 & 17.77 / 15.69   & 5.57 / 27.61   &  15.16 & \textbackslash \\
			Fisher\_\ourapproach (ours)  & 38.87   & 16.98 / 15.44 & 18.85 / 17.36   & 15.46 / \textbf{30.41}  &  16.34 & \textbackslash \\ \midrule
			RegMean          & 39.46   & 21.09 / 14.12 & 18.97 / 16.21   & 15.92 / 4.88  & 20.33 & \textbackslash \\ 
			RegMean\_\ourapproach (ours) & 41.13   & 23.41 / 16.45 & 21.44 / 18.31   & 18.49 / 11.27  & 22.07 & \textbackslash \\ \midrule
			TIES             & 40.48   & 22.63 / 16.34 & 19.76 / 17.52   & 16.88 / 14.72  & 21.92 & 41.35 \\ 
			\textbf{TIES\_\ourapproach} (ours)    & \textbf{42.53 }  & \textbf{24.11 / 19.02} & \textbf{22.64 / 19.06}   & \textbf{18.74} / 16.31  & \textbf{22.75} & \textbf{42.86} \\ \midrule
			MTL      & 37.64  & 27.41      & 25.63        & 31.45       & 25.26  & 44.62   \\ 
			\bottomrule
		\end{tabular}
	}
	\caption{\small{Out-of-domain Performance. All the results we reported are averages of trials conducted with 5 different random seeds.}}
	\label{tab:merge_all_ood_emotion}
\end{table*}


\subsection{Mutation and Crossover}  
\begin{figure}[t]
\vspace{-0.3cm}
\centering
\captionsetup[subfigure]{justification=centering}
        \begin{subfigure}{0.22\textwidth}
            \centering
            \includegraphics[width=\textwidth]{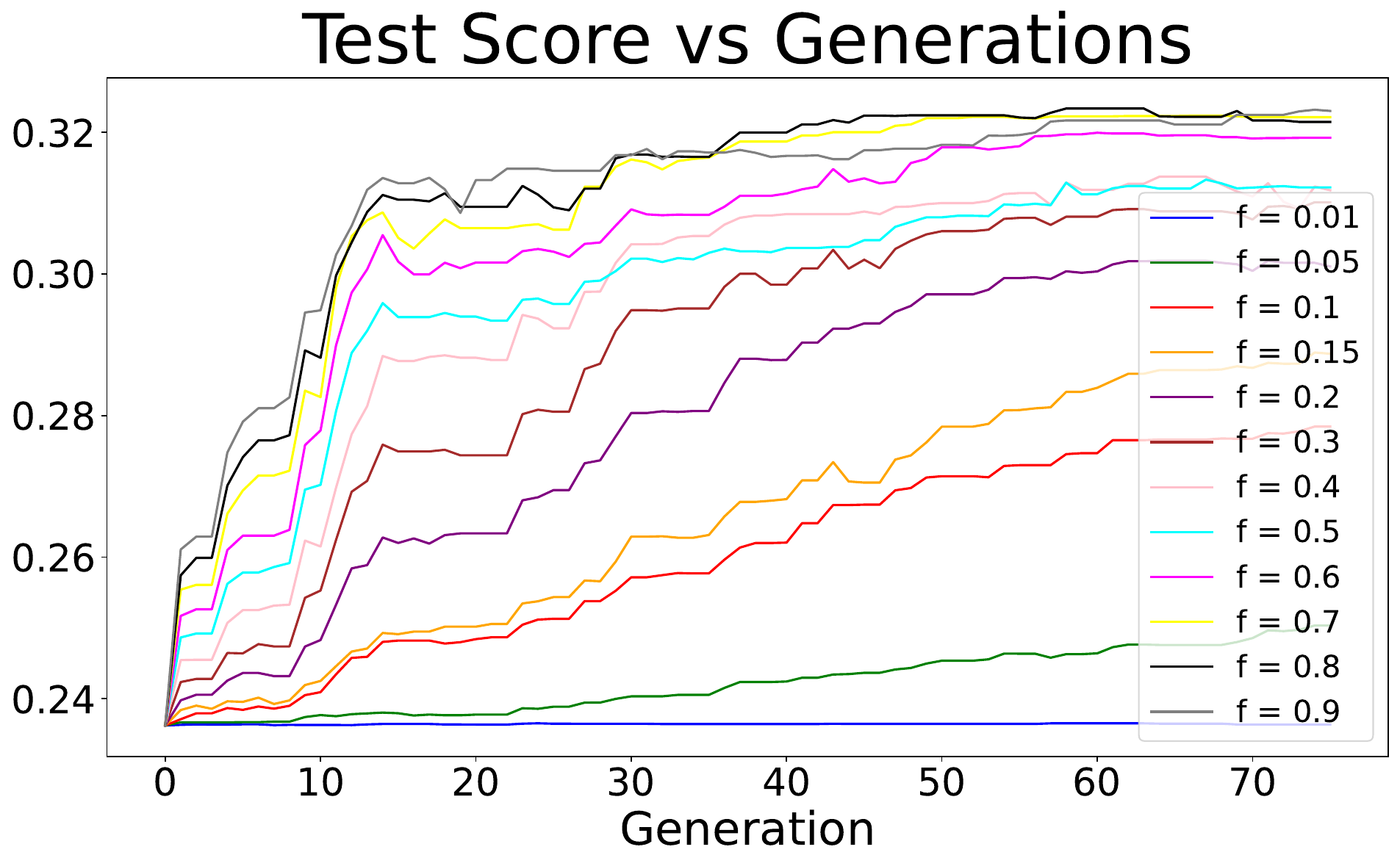}
            \caption{\small{evolver with different \textit{F} when \textit{Cr}=0.5}.}
            \label{fig:evolver-f}
        \end{subfigure}
        \begin{subfigure}{0.22\textwidth}
            \centering
            \includegraphics[width=\textwidth]{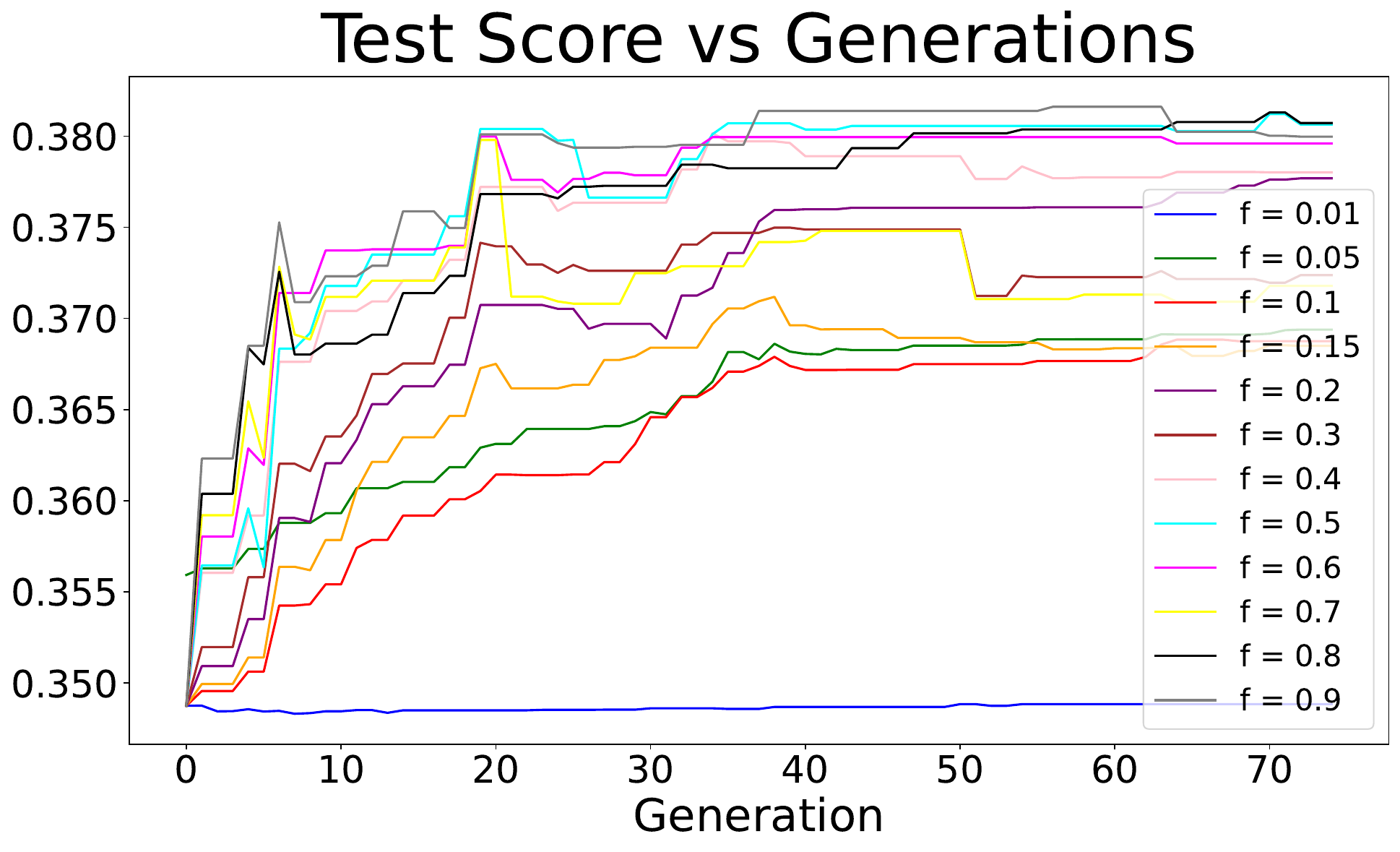}
            \caption{\small{regmean-evolver with different \textit{F} when \textit{Cr}=0.5}.}
            \label{fig:reg-evolver-f}
        \end{subfigure}
        \label{fig:iteration}
        \begin{subfigure}{0.22\textwidth}
            \centering
            \includegraphics[width=\textwidth]{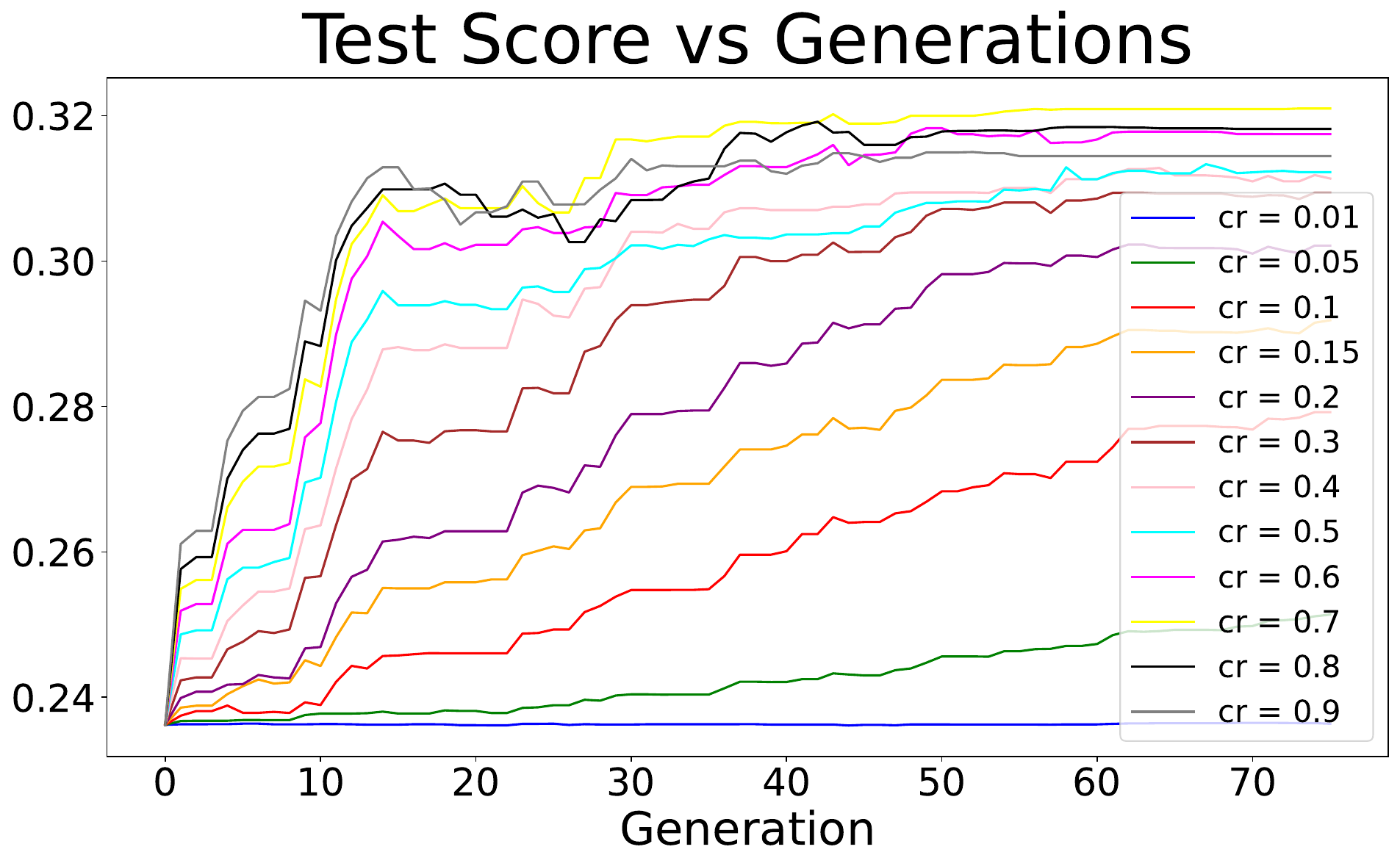}
            \caption{\small{evolver with different \textit{Cr} when \textit{F}=0.5}.}
            \label{fig:evolver-cr}
        \end{subfigure}
        \begin{subfigure}{0.22\textwidth}
            \centering
            \includegraphics[width=\textwidth]{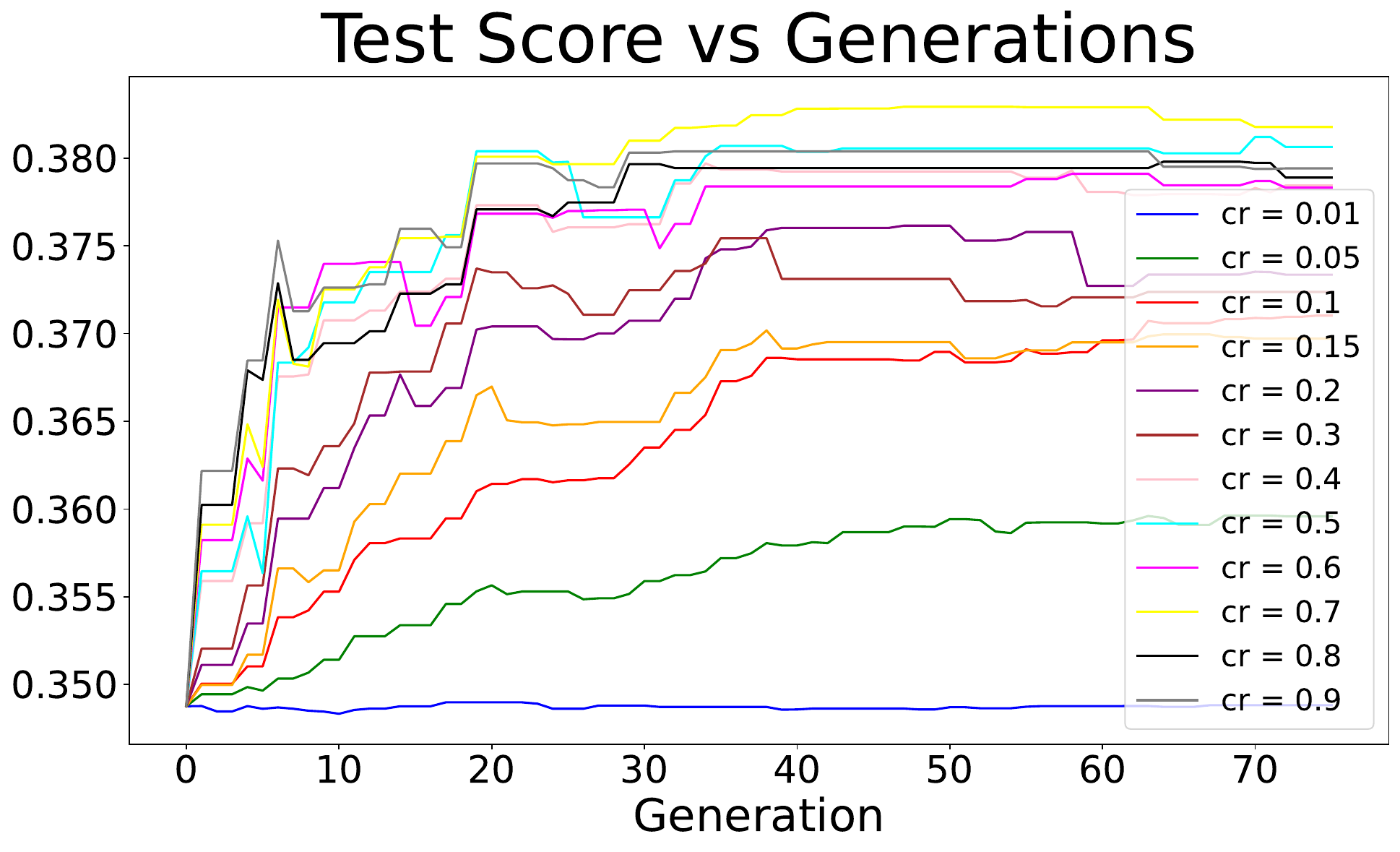}
            \caption{\small{regmean-evolver with different \textit{Cr} when \textit{F}=0.5}.}
            \label{fig:reg-evolver-cr}
        \end{subfigure}
        \caption{Result of RoBERTa-base when evolving all domain specific models on emotion datasets with different with different scale factor \textit{F} and crossover ratio \textit{Cr}.}
        \label{fig:ablation f and cr}
        \vspace{-0.3cm}
\end{figure}

We also test the impact of different values of scale factor \textit{F} for mutation and crossover ratio \textit{Cr}, as shown in Figure~\ref{fig:ablation f and cr}. Due to the inherent randomness in the search process of evolutionary algorithms, we conducted each experiment using four different random seeds and then calculated the average results.
In general, the study findings indicate that the performance of the evolutionary algorithm improves as the parameters \textit{F} or \textit{Cr} increase until reaching 0.5. However, when \textit{F} or \textit{Cr} exceeds 0.5, there is minimal improvement in performance, and no clear pattern is observed. Therefore, in all experiments conducted in this paper, we have consistently used $F=0.5$ and $Cr=0.5$. 

\begin{figure}[t]
	\hfill
	\begin{minipage}{0.48\textwidth} 
		\centering
		\includegraphics[width=\textwidth]{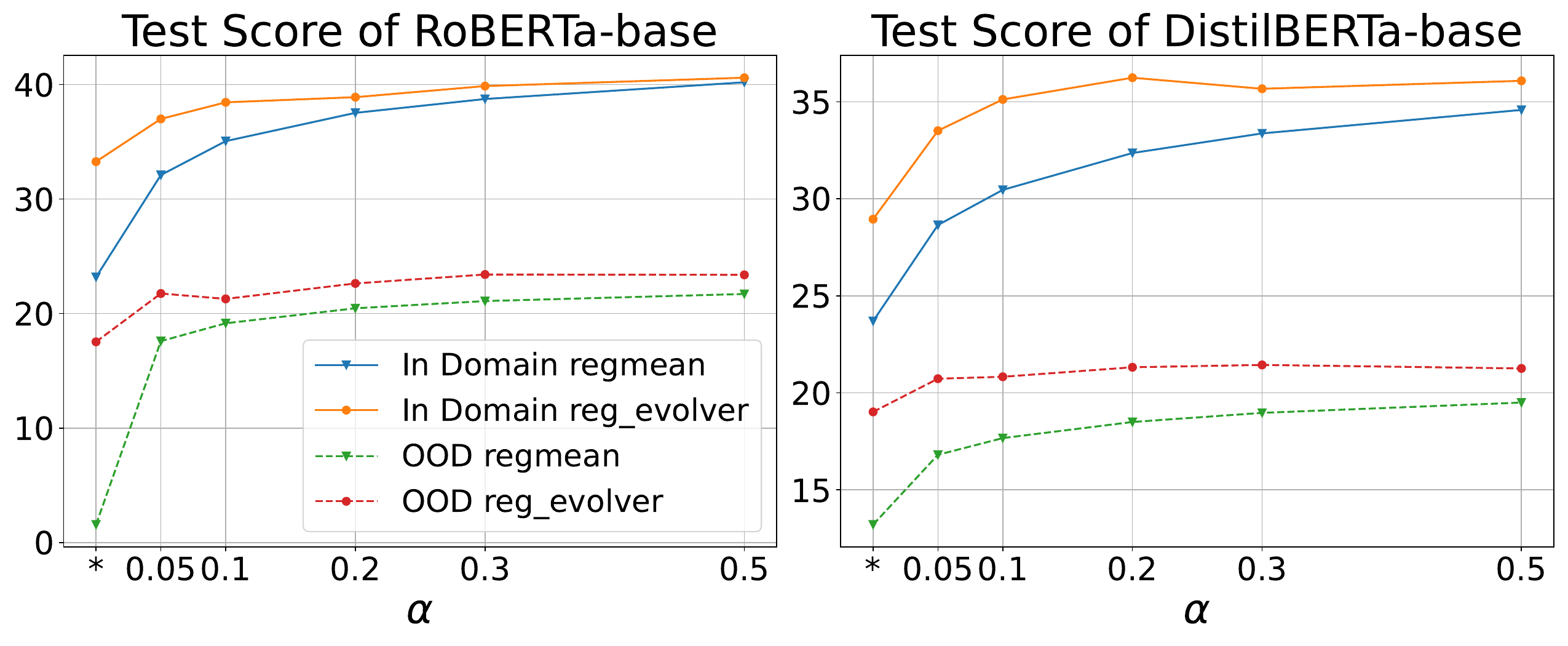}
		\caption{The improvement of \textit{regmean-evolver} with different scale $\alpha$ on the emotion dataset when evolving all domain specific models. $*$ denotes results of simple \textit{evolver}.}
		\label{fig:diff_afa}
	\end{minipage}
\end{figure}

In addition, \textit{regmean} method requires decreasing the non-diagonal items of the inner product matrices by multiplying a scalar $\alpha$. Since the effectiveness of the \textit{regmean-evolver} method can be influenced by the hyperparameter $\alpha$, we also test the performance of \textit{regmean-evolver} under different $\alpha$ parameters, as shown in Figure~\ref{fig:diff_afa}.

\begin{table*}[t]
\centering
\scalebox{0.64}{
\begin{tabular}{@{}c|cc|cc|cc@{}}
\toprule
Model & \begin{tabular}[c]{@{}c@{}}Simple \\ (coefficient search)\end{tabular} 
& \begin{tabular}[c]{@{}c@{}}Evolver \\ (scale factor search)\end{tabular}
& \begin{tabular}[c]{@{}c@{}}Fisher \\ (coefficient search)\end{tabular} 
& \begin{tabular}[c]{@{}c@{}}Fisher\_Evolver \\ (scale factor search)\end{tabular}
& \begin{tabular}[c]{@{}c@{}}RegMean \\ (coefficient search)\end{tabular} 
& \begin{tabular}[c]{@{}c@{}}RegMean\_Evolver \\ (scale factor search)\end{tabular} \\
\cmidrule(r){1-1} \cmidrule(lr){2-3} \cmidrule(lr){4-5} \cmidrule(lr){6-7} 

RoBERTa-base     & 37.78  (\textbf{38.83})  & 39.13  (\textbf{39.98}) & 37.11  (\textbf{38.96)} & 40.34  (\textbf{41.23}) & 46.56  (\textbf{46.82}) & 46.89  (\textbf{47.03}) \\
DistillBERT-base & 36.76  (\textbf{37.63})  & 38.85  (\textbf{39.67}) & 34.52  (\textbf{37.54}) & 40.37  (\textbf{41.31}) & 43.09  (\textbf{43.14}) & 43.22  (\textbf{43.31}) \\
T5-base          & 38.82  (\textbf{39.91})  & 40.21  (\textbf{41.11}) & 38.08  (\textbf{39.22}) & 41.46  (\textbf{42.55}) & 47.35  (\textbf{47.84}) & 47.92  (\textbf{48.06}) \\
\bottomrule
\end{tabular}
}
\caption{\small{\textbf{Coefficient Search Result} when merging pairwise emotion classification models. Simple, Fisher and RegMean are model merging algorithms for comparison. All the results we reported are averages of 10 ($\mathcal{C}^2_5$) runs after paring models from a set of 5.}}
\label{tab:coefficient_search}
\vspace{-0.3cm}
\end{table*}


\subsection{Integration with Coefficient Search}
The coefficient search is a promising scheme to improve the model merging performance, by searching the optimal $\alpha$. The proposed model evolution can also be integrated with the coefficient search method by searching the optimal scale factor $f$. We have performed the grid search of $\alpha$ and $f$ with intervals of 0.05 from 0.1 to 0.9. We present the result in Table \ref{tab:coefficient_search}.
In the setting of Simple, Fisher and RegMean, the results show that a default version of evolver (with scale factor $f=0.5, cr=0.5$) outperforms the coefficient search results. Notably, the integration with scale factor search further boosts the performance of the evolver, which is worth further investigation. Especially, the crossover ratio $Cr$ in model evolution could also be the subject of coefficient search. Many adaptive schemes are also available in the realm of evolution algorithms, like SADE~\citep{qin2005self} and SHADE~\citep{tanabe2013success}, providing a possibility of future algorithmic development. 

\subsection{Analysis}
\begin{figure}[htbp]
	\begin{minipage}{0.48\textwidth} 
		\centering
		\includegraphics[width=\textwidth]{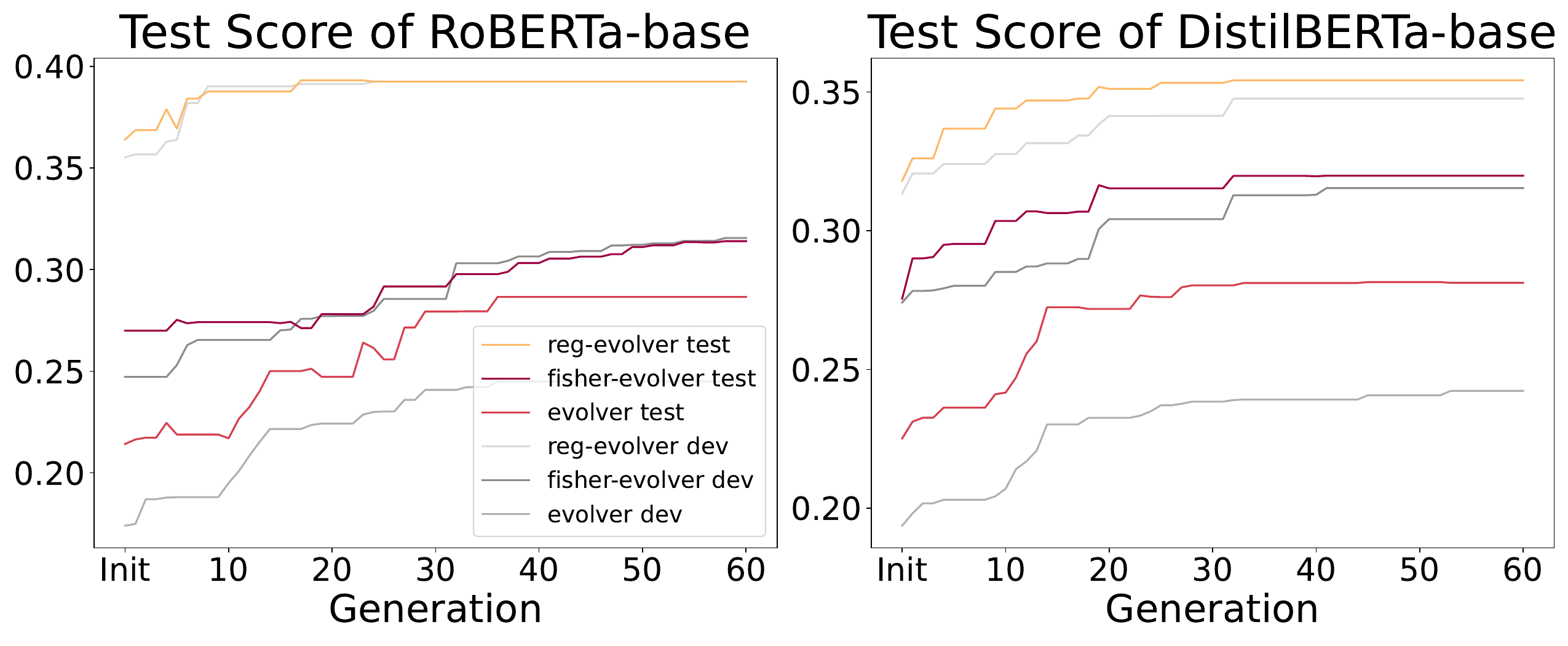}
		\caption{In-domain score of model evolution on the emotion dataset with all domain specific models.}
		\label{fig:combined}
	\end{minipage}
\end{figure}

We demonstrate the evolutionary process when combined with other model fusion methods. From Figure~\ref{fig:combined}, we can observe that when model evolver is combined with \textit{fisher} or \textit{regmean} method, the upper bounds of the evolutionary approach can be enhanced. Additionally, we present the test results of the evolutionary algorithm on the development dataset. It is evident that as individual models are trained on the development dataset, their performance on the test set gradually improves. This indicates that our model evolution method indeed has the ability to optimize and learn.

\begin{figure}[htbp]
	\centering
	\includegraphics[width=0.5\textwidth]{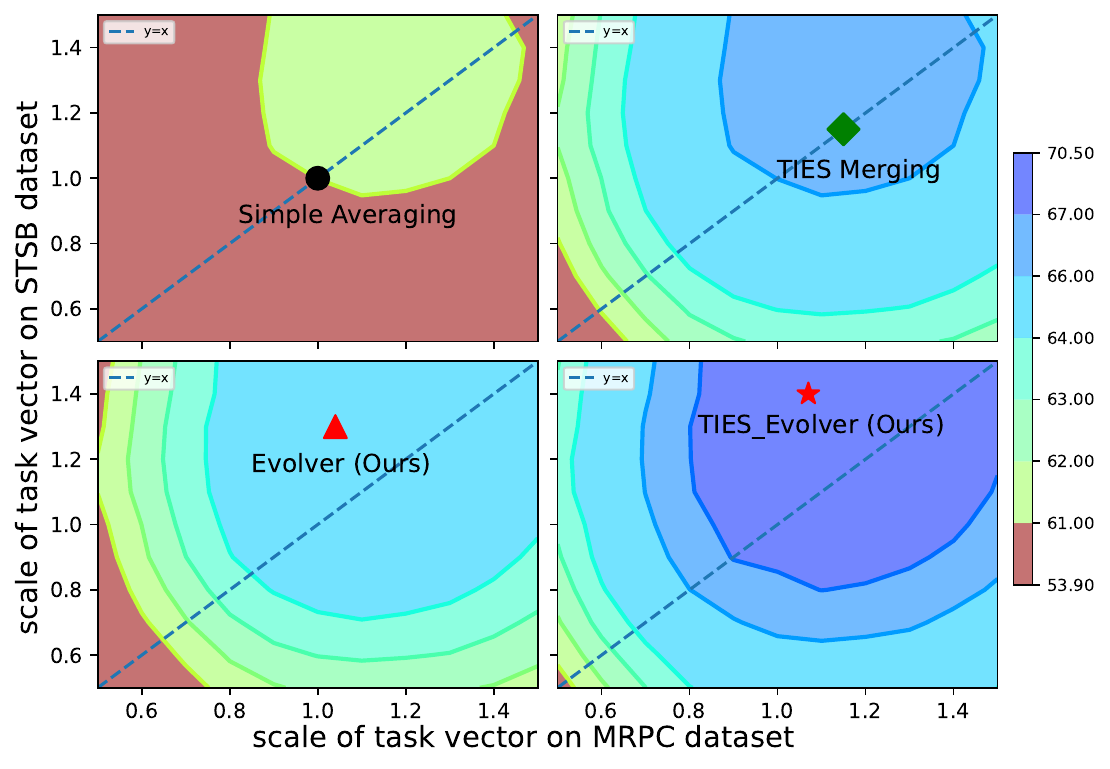}
	\caption{The average performance of merged T5 model on MRPC and STSB test dataset. By observing the density and distribution range of the contour lines, it is evident that our evolver method performs better. Additionally, previous methods typically utilize a uniform scale across different task vectors, whereas the evolved model exhibits a preference for varying scales determined through iterative evolution rounds.}
	\label{fig:hotmap}
\end{figure}

In addition, we visualize a two dimensional slice of the average test accuracy landscape when evolving pairwise fine-tuned T5 models, as shown in Figure~\ref{fig:hotmap}. In this experiment, we use the zero-shot initialization and fine-tune twice, independently, to produce task vectors on MRPC and STSB datasets. We analyze the positions of the model results after fusion by different methods in their corresponding landscapes. Methods like TIES and other merging approaches typically process the task vectors of models on different tasks and then perform a direct weighted average. This results in a fixed relative scale among different tasks during model fusion. However, from the landscape, we can see that the optimal fusion performance usually has different preferences for different tasks. Our proposed model evolution method can dynamically update the scales of different tasks to promote better model fusion effects.

The proposed model evolution also has some advantages in other aspects:
(1) Model evolution can leverage the benefits of a larger population size without being significantly affected by individuals with extremely poor performance. This advantage is a result of the survival of the fittest mechanism in the model evolution process. By favoring those with most effective performance, we reduce the influence of individuals with extreme differences, leading to a more robust and reliable system.
(2) Model evolution method can effectively maintain low peak GPU memory usage. This is primarily attributed to its sequential forward inference of individual models, as opposed to most previous model merging techniques that require additional GPU memory for computing inner product matrices in the model parameter space. This advantage significantly reduces GPU memory consumption and extends the range of feasible solutions for large-scale language models.

\section{Conclusions}

We introduce a novel knowledge fusion method, called model evolution, inspired by evolutionary algorithms. This approach significantly boosts the performance of model merging in diverse NLP contexts. Model evolution stands out by aggregating model weights into a population and updating it with superior offspring models, all without requiring extra training data. The most significant contribution of our work lies in our experimental discovery that knowledge fusion can benefit from searching per-model and per-parameter coefficients. This type of search-based, gradient-free optimization algorithm proves to be an effective tool for model merging and warrants greater attention in the research community. Our extensive experiments validate its superiority over previous techniques.
\clearpage

\section*{Limitation}
The limitations of the model evolution method can be summarized in three main aspects: (1) the necessity for a high-quality development dataset with consideration for data privacy, (2) the requirement for a cautious selection of hyperparameters \textit{F} and \textit{Cr}, (3) and the significance of conducting further theoretical analysis of evolutionary algorithm principles.
Future research offers several promising directions. Firstly, exploring advanced optimization strategies within evolutionary algorithms, including adaptive approaches and hyperparameter selection based on historical performance, holds the potential for enhancing the method's effectiveness. Secondly, extending knowledge fusion to a more complex training environment by considering hyperparameters, multimodal and exploring different training methods like unsupervised or supervised learning can provide a comprehensive understanding of its applicability. Thirdly, arithmetic operations in \cite{ilharco2022editing} for model edit can be analyzed. Lastly, evaluating the approach on larger language models can provide insights into its scalability.
\section*{Ethical Considerations}
Our approach has been evaluated on GLUE. We explicitly claim that the applicability of our method and findings may be confined to similar datasets or domains. The performance of our method on other specific datasets or domains remains uncertain. Thus, there are potential risks when applying our method to privacy-sensitive or high-risk datasets. We should be cautious and verify whether the method generates correct answers.
\section*{Acknowledgements}
This work was supported in part by Shenzhen College Stability Support Plan (GXWD20231128103232001), Department of Science and Technology of Guangdong (2024A1515011540), Shenzhen Start-Up Research Funds~(HA11409065), National Natural Science Foundation of China (12204130), the Ministry of Higher Education Malaysia through the Fundamental Research Grant Scheme (FRGS/1/2023/ICT02/XMU/02/1), and Xiamen University Malaysia through Xiamen University Malaysia Research Fund (XMUMRF/2022-C10/IECE/0039 and XMUMRF/2024-C13/IECE/0049).

\newpage
\bibliography{custom}
\bibstyle{acl_natbib}

\appendix
\label{sec:appendix}
\newpage
\appendix
\section*{Appendix}
\section{Explanation of Model Evolution}\label{appendix A}
\begin{figure}[h]
  \begin{center}
    \includegraphics[width=0.4\textwidth]{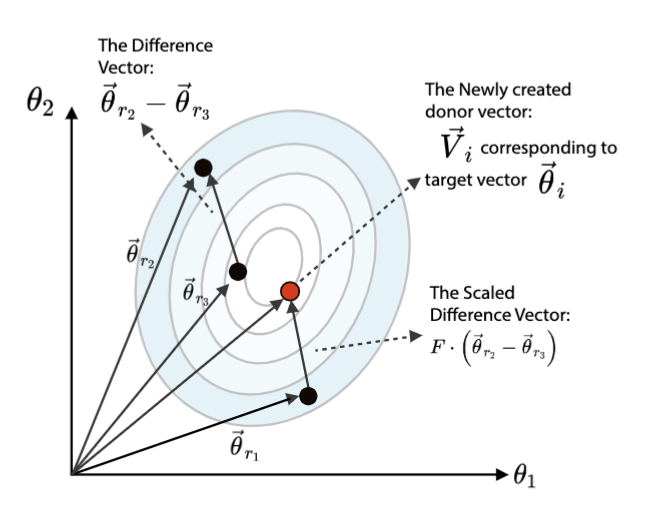}
  \end{center}
\caption{An illustration of the mutation process in difference evolution. }
\label{fig: de-mutation}
\end{figure}

In this section, we provide a more comprehensive explanation of the principles underlying the Differential Evolution algorithm, furthermore, we offer an in-depth elucidation of the mutation process, providing a visual representation in Figure~\ref{fig: de-mutation} to enhance clarity and understanding. Overall, the fundamental principle of the differential evolution algorithm involves randomly selecting three distinct individuals, performing a mutation operation to create a new candidate solution, using a crossover operation to refine the solution, and replacing the original solution if the new one performs better. This iterative process continues until certain stopping criteria are met. 

To better illustrate our model evolution method, we have created a flowchart as shown in algorithm~\ref{flowchart}. The details of combining our proposed method with other models are provided in Step 3. The approach involves calculating an overall score by using model merging on the mutated individuals along with the non-mutated individuals in the current evolution. In contrast, the simple evolver determines the success of mutation based on the score of individual entities, while the combined approach assesses it based on the score of the entire population after individual mutation.
\begin{algorithm*}[htbp]
\caption{Model Evolution}
\begin{algorithmic}[1]
\State \textbf{Step 1 - Initializing the Population} 
\State Initialize population $\Theta$ \Comment{A population of candidate solutions}
\State $generation \gets 0$ \Comment{Initialize generation counter}
\State $converged \gets$ \textbf{False} \Comment{Convergence flag}\\
\textbf{while} not converged \textbf{do}
\State \hspace{1cm} \textbf{Step 2 - Evolution Process: Mutation and Recombination}
\State \hspace{1cm} \textbf{for} each candidate solution 
 $\theta_i$ in $\Theta$ \textbf{do}
\State \hspace{2cm} Randomly select $\theta_{r1}$ and $\theta_{r2}$ \Comment{Select random solutions}
\State \hspace{2cm} $F \gets$ Random scaling factor \Comment{Control parameter for mutation}
\State \hspace{2cm} $Cr \gets$ Random crossover rate \Comment{Control parameter for recombination}
\State \hspace{2cm} Compute mutated solution $\theta^\star_i$ using $\theta_i$, $\theta_{r1}$, $\theta_{r2}$, and $F$
\State \hspace{2cm} Perform recombination of $\theta^\star_i$ based on $Cr$ and $\theta_i$
\State \hspace{2cm} \textbf{Step 3 - Model Inference}
\State \hspace{2cm} \textbf{if} not combined with other model merging methods \textbf{then} 
\State \hspace{3cm} Evaluate the performance of $\theta^\star_i$ on development data
\State \hspace{2cm} \textbf{else} 
\State \hspace{3cm} Merging $\theta^\star_i$ with other models  
\State \hspace{3cm} Evaluate the performance of merged model on development data
\State \hspace{2cm} \textbf{end if}
\State \hspace{1cm} \textbf{end for}
\State \hspace{1cm} \textbf{Step 4 - Updating the Population}

\State \hspace{1cm} $converged \gets$ \textbf{True} \Comment{Assume convergence}
\State \hspace{1cm} \textbf{for} each candidate solution $\theta_i$ in $\Theta$ \textbf{do}
\State \hspace{2cm}\textbf{if} $\theta^\star_i$ outperforms $\theta_i$ \textbf{then} \Comment{Comparing performance}
\State \hspace{3cm}Replace $\theta_i$ with $\theta^\star_i$ \Comment{Update population}
\State \hspace{3cm}$converged \gets$ \textbf{False} \Comment{Reset convergence flag}
\State \hspace{2cm} \textbf{end if}
\State \hspace{1cm} \textbf{end for}
\State \hspace{1cm}$generation \gets generation + 1$ \Comment{Increment generation counter}
\State \textbf{end while}
\end{algorithmic}
\label{flowchart}
\end{algorithm*}
\vspace{-0.3cm}


\section{Preliminaries}

\paragraph{Fisher-weighted averaging (\textit{fisher})} \cite{matena2022merging} examines the importance of each weight $F_i$ associated with each label by computing the norm of the logarithmic likelihood gradient. Specifically, the posterior probabilities of each model are interpreted as gaussian distributions $p(\theta | \theta_i, F_i)$, where the parameters $\theta$ for model $i$ correspond to the Fisher information matrix $F_\theta$. Finally, the fisher information for each parameter is used to perform a weighted average of the parameters, integrating the parameters of different models into a single model.

\paragraph{Regression mean \textit{(regmean)}}\cite{jin2022dataless} expands the solution of a linear optimization problem to $K$ models where $W_i,i \in \mathcal{K}$, denoted as $W_M = (\sum_i^{i \in \mathcal{K}} X_i^TX_i)^{-1} \sum_i^{i \in \mathcal{K}}(X_i^T X_i W_i)$. Each transformer model  $f^{(j)}$ linear layer's corresponding $X_i^{(j)}$ is captured along with per-weights and its input inner product matrix, to compute the merged weights and produced merged model $f_M(x)=W_{M}^{T} x$. The scale of $X_i^T X_i$ exhibits substantial variation across different models. Additionally, a control mechanism is applied by multiplying $X_i^T X_i$  by $\alpha=\frac{1}{1+\gamma}$.

\paragraph{Model soups (\textit{greedy soup})} \cite{wortsman2022model}. Initially, models are ranked based on their development dataset scores. Subsequently, model parameters $\theta_i$ are chosen through a greedy search, and their inclusion in the gradient is determined by comparing the average validation set accuracy before and after their addition. The merged model's parameters can be represented as $\theta_S = \text{average}(\text{ingredients})$.

\paragraph{TIES merging (\textit{TIES})} \citep{yadav2023ties}. 
This method is based on the concept of task vectors \cite{ilharco2022editing} and address the problem of two major sources of interference: (a) interference
due to redundant parameter values and (b) disagreement on the sign of a
given parameter’s values across models. This method proposed TRIM, ELECT SIGN \& MERGE (TIES-MERGING), which introduces three novel
steps when merging models: (1) resetting parameters that only changed a small
amount during fine-tuning, (2) resolving sign conflicts, and (3) merging only the
parameters that are in alignment with the final agreed-upon sign
\section{Impact of Development Dataset}\label{appendix B}
The availability of development datasets directly impacts the effectiveness of our model evolution approach. However, many publicly available datasets either do not provide development sets or widely use them as test sets. In our case, the development set of the GLUE dataset is used as a test set, so we utilize a small portion of the training dataset (approximately 5\%) for model evolution. For non-i.i.d. partition methods, we also use only a subset of the same training data samples. In the case of the unified emotion dataset, we separately extract 10\% of data from each of the five high-resource datasets for model evolution, following the same partitioning method as employed in ~\cite{jin2022dataless}.

For our model evolution approach, the quality of the development dataset can significantly impact performance, making the selection of a high-quality development dataset a crucial consideration. To address this, we conducted experiments on the emotion dataset using different lengths of model evolution methods. We performed experiments with both the simple evolver and regmean evolver, evolving all five domain-specific models. The experimental results are shown in Table~\ref{tab: length}, indicating that even with a short development dataset, model evolution can still be effective. However, as the length of the development dataset increases, the performance of model evolution tends to improve. Additionally, we included the test scores of simple methods as baselines for comparison.
\begin{table}[h]
\centering
\scalebox{0.9}{
\begin{tabular}{cccccc}
\toprule
Length & None & 1/4 & 1/2 & 1 \\
\midrule
Evolver & 23.18 & 30.14 & 32.03 & 33.27 \\
\midrule
Regmean\_Evolver & 38.74 & 39.43 & 39.57 & 39.87 \\
\bottomrule
\end{tabular}}
\caption{The performance of model evolution with different length of development dataset. \textit{None} means evolver is not conduct and the test score of \textit{simple averaging} and \textit{regmean} method is recorded.}
\label{tab: length}
\end{table}

\section{Dataset, Metrics and Training Details}\label{appendix C}
\subsection{Emotion Classification Datasets}
In order to investigate the performance of the sentiment classification task, we selected a diverse and challenging set of datasets. Among them, DailyDialogs~\citep{li2017dailydialog}, CrowdFlower, TEC~\citep{mohammad2012emotional}, Tales-Emotion~\citep{alm2005emotions}, and ISEAR~\citep{scherer1994evidence} is utilized to train domain-specific model. For acessing OOD generalization performance, we use Emoint~\citep{mohammad2017wassa}, SSEC~\citep{schuff2017annotation}, ElectoralTweets~\citep{mohammad2015sentiment}, GroundedEmotions~\citep{liu2017grounded}, and AffectiveText~\citep{strapparava2007semeval}. For OOD evaluation, we focus exclusively on the fundamental emotions: anger, disgust, fear, joy, sadness, and surprise. 
A detailed overview of the datasets and statistics is provided in Table~\ref{tab:data_stats_emotion}.

\subsection{GLUE Benchmark Datasets}
For the GLUE dataset, we utilized multiple tasks, including CoLA~\citep{warstadt2019neural}, SST-2~\citep{socher2013recursive}, MRPC~\citep{dolan2005automatically}, STS-B~\citep{cer2017semeval}, MNLI~\citep{williams2018broad},QNLI~\citep{rajpurkar2016squad}, QQP, and RTE~\citep{giampiccolo2007third}. These tasks cover various natural language understanding problems such as text classification, text similarity, and natural language inference. To assess our merged models, we tested them on the official development sets. We performed experiments by training models on non-i.i.d. partitions, creating various partition scenarios through random sampling. Each partition is uniformly sub-sampled to yield a total of 1,000 training examples per partition.

\subsection{Definitions and Metrics} \label{app_c.2}
\begin{table}
\centering
\scalebox{0.9}{
\begin{tabular}{@{}lrrr@{}}
\toprule
                   & Train  & Dev    & Test   \\ \midrule
\textit{In-domain} &        &        &        \\
DialyDialog        & 72,085 & 10,298 & 20,596 \\
CrowdFlower        & 27,818 & 3,974  & 7,948  \\
TEC                & 14,735 & 2,105  & 4,211  \\
Tales-Emotion      & 10,339 & 1,477  & 2,955  \\
ISEAR              & 5,366  & 766    & 1,534  \\ \midrule
\textit{Out-of-domain}       &        &        &        \\ 
Emoint             &        &        & 7,102  \\
SSEC               &        &        & 4,868  \\
ElectoralTweets    &        &        & 4,056  \\
GroundedEmotions   &        &        & 2,585  \\
AffectiveText      &        &        & 1,250 \\
\bottomrule
\end{tabular}
}
\caption{\small{Statistics of emotion classification datasets.}}
\label{tab:data_stats_emotion}
\end{table}

\begin{table*}[htbp]
\centering
\scalebox{0.72}{
\begin{tabular}{@{}lcccccc@{}}
\toprule
Initial Population for Evolving & T5-base & RoBERTa-base & DistilBERT-base  & DeBERTa-large & GPT2 & MiniCPM \\
\cmidrule(r){1-1} \cmidrule(lr){2-2} \cmidrule(lr){3-5} \cmidrule(lr){6-7} 
All Domain Specifis Models on Emotion Datasets & 24.5 & 19.2 & 18.7 & 21.3 & 20.1 & 28.1 \\
Pairwise Models on Emotion Datasets            & 9.3 & 7.3 & 7.1 & 8.1 & 7.6 & 13.2\\
None-iid Pairwise Models on GLUE Benchmark     & 7.2 & 5.7  & 5.5 & 6.2 & 5.9 & 10.6\\ 
Cross Tasks Pairwise Models on GLUE Benchmark  & 8.7 & 7.1 & 6.8 & 7.8  & 7.5 & 12.6\\ \midrule
\bottomrule
\end{tabular}
}
\caption{\small{\textbf{Time cost} (in the unit of minutes) of RegMean\_Evolver on different experiments with 20 generations. T5-base is tested on single A800 GPU and other models are tested on single A6000 GPU. The time cost is mainly related to the size of model and the length of development dataset when conducting model evolution.}}
\label{tab:time_cost}
\end{table*}

The performance of individual models involved in merging are reported: (1) the average performance of all individual models (\textbf{Avg. $f_{1..N}$}); (2) the performance of the best \textit{single} individual model (\textbf{Best. $f_{1..N}$}), as determined by using the validation set; (3) the performance of the individual models corresponding to the training data set for each test set (\textbf{Domain-Specific}). 

In evaluating merged models trained for non-i.i.d. partitions of the same dataset, we assessed their performance using a unified test set characterized by a joint distribution of all partitions. For merged models trained across different domains or tasks, we measured their performance across individual domains or tasks incorporated into the merger and derived their macro-average. Similarly, when evaluating out-of-domain performance, we computed the macro-average of their performance across the out-of-domain test set.

\subsection{Merging Models Trained on Non-i.i.d. Partitions.} \label{app_c.1}
Merging models initially trained on non-i.i.d. partitions of the same dataset is started, which is achieved by simulating synthetic data splits across the 8 tasks within the GLUE benchmark. Each task involves dividing the training data into two partitions, each containing 1,000 training examples with distinct label distributions. Following this, we perform fine-tuning on these two partitions for 8 pairs of individual models and merge each pair of models. The evaluation of these merged models takes place on the official validation sets, which portray a joint distribution of both partitions.

\subsection{Time Cost}
We report the time cost of the scheme of model evolution, as shown in Table~\ref{tab:time_cost}. T5-base model and MiniCPM model are tested on single NVIDIA A800 80G GPU and other models are tested on single RTX A6000 48G GPU. We find that all task of model evolution can be completed within half an hour, which is very cost-efficient in improving the model performance without further training.

\end{document}